# A Comparison of WordNet and Roget's Taxonomy for Measuring Semantic Similarity

Michael L. Mc Hale
Intelligent Information Systems
Air Force Research Laboratory
525 Brooks Road
13441 Rome, NY, USA,
mchale@ai.rl.af.mil

## Abstract

This paper presents the results of using Roget's International Thesaurus as the taxonomy in a semantic similarity measurement task. Four similarity metrics were taken from the literature and applied to Roget's. The experimental evaluation suggests that the traditional edge counting approach does surprisingly well (a correlation of r=0.88 with a benchmark set of human similarity judgements, with an upper bound of r=0.90 for human subjects performing the same task.)

## Introduction

The study of semantic relatedness has been a part of artificial intelligence and psychology for many years. Much of the early semantic relatedness work in natural language processing centered around the use of Roget's thesaurus (Yaworsky 92). As WordNet (Miller 90) became available, most of the new work used it (Agirre & Rigau 96, Resnik 95, Jiang & Conrath 97). This is understandable, as WordNet is freely available, fairly large and was designed for computing. Roget's remains, though, an attractive lexical resource for those with access to it. Its wide, shallow hierarchy is densely populated with nearly 200,000 words and phrases. The relationships among the words are also much richer than WordNet's IS-A or HAS-PART links. The price paid for this richness is a somewhat unwieldy tool with ambiguous links.

This paper presents an evaluation of Roget's for the task of measuring semantic similarity. This is done by using four metrics of semantic similarity found in the literature while using Roget's International Thesaurus, third edition (Roget 1962) as the taxonomy. Thus the results can be compared to those in the literature (that used WordNet). The end result is the ability to compare the relative usefulness of Roget's and WordNet for this type of task.

## 1 Semantic Similarity

Each metric of semantic similarity makes assumptions about the taxonomy in which it works. Generally, these assumptions go unstated but since they are important for the understanding of the results we obtain, we will cover them for each metric. All the metrics assume a taxonomy with some semantic order.

### 1.1 Distance Based Similarity

A common method of measuring semantic similarity is to consider the taxonomy as a tree, or lattice, in semantic space. The distance between concepts within that space is then taken as a measurement of the semantic similarity.

#### 1.1.1 Edges as distance

If all the edges (branches of the tree) are of equal length, then the number of intervening edges is a measure of the distance. The measurement usually used (Rada et al. 89) is the shortest path between concepts. This, of course, relies on an ideal taxonomy with edges of equal length. In taxonomies based on natural languages, the edges are not the same length. In Roget's, for example, the distance (counting edges) between *Intellect* and *Grammar* is the same as the distance between *Grammar* and *Phrase Structure*. This does not seem intuitive. In general, the edges in this type of taxonomy tend to grow shorter with depth.

### 1.1.2 Related Metrics

A number of different metrics related to distance have used edges that have been modified to correct for the problem of non-uniformity. The modifications include the density of the subhierarchies, the depth in the hierarchy where the word is found, the type of links, and the information content of the nodes subsuming the word.

The use of density is based on the observation that words in a more densely part of the hierarchy are more closely related than words in sparser areas (Agirre and Rigau 96). For density to be a valid metric, the hierarchy must be fairly complete or at least the distribution of words in the hierarchy has to closely reflect the distribution of words in the language. Neither of these conditions ever hold completely. Furthermore, the observation about density may be an overgeneralization. In Roget's, for instance, category **277 Ship/Boat** has many more words (much denser) than category **372 Blueness**. That does not mean that *kayak* is more closely related to *tugboat* than *sky blue* is to *turquoise*. In fact, it does not even mean that *kayak* is closer to **Ship/Boat** than *turquoise* is to **Blueness**.

Depth in the hierarchy is another attribute often used. It may be more useful in the deep hierarchy of WordNet than it is in Roget's where the hierarchy is fairly flat and uniform. All the words in Roget's are at either level 6 or 7 in the hierarchy.

The type of link in WordNet is explicit, in Roget's it is never clear but it consists of more than IS-A and HAS-PART. One such link is HAS-ATTRIBUTE.

Some of the researchers that have used the above metrics include Sussna (Sussna 93) who weighted the edges by using the density of the subhierarchy, the depth in the hierarchy and the type of link. Richardson and Smeaton (Richardson and Smeaton 95) used density, hierarchy depth and the information content of the concepts. Jiang and Conrath (Jiang and Conrath 95) used the number of edges and information content. They all reported improvement in results compared to straight edge counting.

McHale (95) decomposed Roget's taxonomy and used five different metrics to show the usefulness of the various attributes of the taxonomy. Two of those metrics deal with distance but only one is of interest to us for this task; the number of intervening words. The number of intervening words ignores the hierarchy completely, treating it as a flat file. For the measurement to be an accurate metric, two conditions must be met. First, the ordering of the words must be correct. Second, either all the words of the language must be represented (virtually impossible) or they must be evenly distributed throughout the hierarchy[1]. Since it is unlikely that either of these conditions hold for any taxonomy, the most that can be expected of this measurement is that it might provide a reasonable approximation of the distance (similar to density). It is included here, not because the approximation is reasonable, but because it provides information that helps explain the other results.

### 1.2 Information Based Similarity

Given the above problems with distance related measures, Resnik (Resnik 95) decided to use just the information content of the concepts and compared the results to edge counting and human replication of the same task. Resnik defines the similarity of two concepts as the maximum of the Information Content of the concepts that subsume them in the taxonomy. The Information Content of a concept relies on the probability of encountering an instance of the concept. To compute this probability, Resnik used the relative frequency of occurrence of each word in the Brown Corpus[2]. The probabilities thus found should fairly well approximate the true values for other generalized texts. The concept probabilities were then computed from the occurrences as simply the relative frequency of the concept.

---

[1] This condition certainly does not hold true in WordNet where animals and plants represent a disproportionately large section of the hierarchy.

[2] Resnik used the semantic concordance (semcor) that comes with WordNet. Semcor is derived from a hand-tagged subset of the Brown Corpus. His calculations were done using WordNet 1.5.

$$Þ(c) = \frac{Freq(c)}{N}$$

The information content of each concept is then given by $IC(c) = \log^{-1} Þ(c)$, where $Þ(c)$ is the probability. Thus, more common words have lower information content.

To replicate the metric using Roget's, the frequency of occurrence of the words found in the Brown Corpus was divided by the total number of occurrences of the word in Roget's[3]. From the information content of each concept, the information content for each node in the Roget hierarchy was computed. These are simply the minimum of the information content of all the words beneath the node in the taxonomy. Therefore, the information content of a parent node is never greater than any of its children.

The metric of relatedness for two words according to Resnik is the information content of the lowest common ancestor for any of the word senses. What this implies is that, for the purpose of measuring relatedness, each synset in WordNet or each semicolon group in Roget's would have an information content equal to its most common member. For example, the words *druid* (Roget's Index number 1036.15) and *pope* (1036.8) would have an information content equal to that of *clergy* (1036). *Clergy*'s information content is based on the two most common words below it in the hierarchy – *brother* and *sister*. Thus *druid* would have an information content less than that of *brother*, a situation that I do not find intuitive since *druid* appears much less frequently than *brother*.

Computationally, the easiest way to compute the information content of a word is to completely compute the values for the entire hierarchy *a priori*. This involves approximately 300,000 (200,000 words plus 100,000 nodes in the hierarchy) computations for the entire Roget hierarchy. This is sizeable overhead compared to edge counting which requires no *a priori* computations. Of course, once the computations are done they do not need to be recomputed until a new word is added to the hierarchy. Since the values for information content *bubble up* from the words, each addition of a word would require that all the hierarchy above it be recomputed.

Jiang and Conrath (Jiang and Conrath 97) also used information content to measure semantic relatedness but they combined it with edge counting using a formula that also took into consideration local density, node depth and link type. They optimized the formula by using two parameters, $\alpha$ and $\beta$, that controlled the degree of how much the node depth and density factors contributed to the edge weighting computation. If $\alpha=0$ and $\beta=1$, then their formula for the distance between two concepts $c_1$ and $c_2$ simplifies to

$$Dist(c_1,c_2) = IC(c_1) + IC(c_2) - 2 \times IC(LS(c_1,c_2))$$

Where $LS(c_1,c_2)$ denotes the lowest super-ordinate of $c_1$ and $c_2$.

## 2  Evaluation

The above metrics are used to rate the similarity of a set of word pairs. The results are evaluated by comparing them to a rating produced by human subjects. Miller and Charles (1991) gave a group of students thirty word pairs and asked the students to rate them for "similarity in meaning" on a scale from 0 (no similarity) to 4 (perfect synonymy). Resnik (1995) replicated the task with a different set of students and found a correlation between the two ratings of $r=.9011$ for the 28 word pairs tested. Resnik, Jiang and Conrath (1997) and I all consider this value to be a reasonable upper-bound to what one should expect from a computational method performing the same task.

Resnik also performed an evaluation of two computational methods both using WordNet 1.5. He evaluated simple edge counting ($r=.6645$) and information content ($r=.7911$). Jiang and Conrath improved on that some ($r=.8282$) using a version of their combined formula given above

---

[3] The frequencies were computed for Roget's as the total frequency for each word divided by the number of senses in Roget. This gives us an approximation of the information content for each concept. The frequency data were taken from the MRC Psycholinguistic database available from the Oxford Text Archive.

that had been empirically optimized for WordNet.

Table 1 gives the results from Resnik (the first four columns) along with the ratings of semantic similarity for each word pair using information content, the number of edges, the number of intervening words and Jiang and Conrath's simplified formula ($\alpha=0$, $\beta=1$) with respect to Roget's. Both the number of edges and the number of intervening words are given in their raw form. The correlation value for the edges was computed using (12 – Edges) where 12 is the maximum number of edges. The correlation for intervening words was computed using (199,427 – words).

## 3  Synopsis of Results

| Similarity Method | Correlation |
| --- | --- |
| **WordNet** | |
|     Human judgements (replication) | r=.9015 |
|     Information Content | r=.7911 |
|     Edge Counting | r=.6645 |
|     Jiang & Conrath | r=.8282 |
| **Roget's** | |
|     Information Content | r=.7900 |
|     Edge Counting | r=.8862 |
|     Intervening Words | r=.5734 |
|     Jiang & Conrath | r=.7911 |

## 4  Discussion

Information Content is very consistent between the two hierarchies. Resnik's correlation for WordNet was 0.7911 while the one conducted here for Roget's was 0.7900. This is remarkable in that the IC values for Roget's used the average number of occurrences for all the senses of the words whereas for WordNet the number of occurrences of the actual sense of the word was used. This may be explainable by realizing that in either case the numbers are just approximations of what the real values would be for any particular text.

Jiang & Conrath's metric did just a little worse using Roget's than the results they gave using WordNet but that may very well be because I was unable to optimize the values of $\alpha$ and $\beta$ for Roget's.

The harder result to explain seems to be edge counting. It does much better in the shallow, uniform hierarchy of Roget's than it does in WordNet. Why this is the case requires further investigation. Factors to consider include the uniformity of edges, the maximum number of edges in each hierarchy and the general organization of the two hierarchies. I expect that major factors are the fairly uniform nature of Roget's hierarchy and the broader set of semantic relations allowed in Roget's. Currently, it seems that Roget's captures the popular similarity of isolated word pairs better than WordNet does.

## 5  Related Work

Agirre and Rigau (Agirre and Rigau 1996) use a conceptual distance formula that was created to by sensitive to the length of the shortest path that connects the concepts involved, the depth of the hierarchy and the density of concepts in the hierarchy. Their work was designed for measuring words in context and is not directly applicable to the isolated word pair measurements done here. Agirre and Rigau feel that concepts in a dense part of the hierarchy are relatively closer than those in a more sparse region; a point which was covered above. To measure the distance, they use a conceptual density formula. The Conceptual Density of a concept, as they define it, is the ratio of areas; the area expected beneath the concept divided by the area actually beneath it.

Some of the results given in Table 1 seem to support the use of density. The word pairs *forest-graveyard* and *chord-smile* both have an edge distance of 8. The number of intervening words for each pair are considerably different (296 and 3253 respectively). For these particular word pairs the latter numbers more closely match the ranking given by humans. If one considers density important then perhaps we can use a different measure of density by computing the number of intervening words per edge[4]. This metric was tested with the 28 word pairs and the results were a slight improvement (r=.6472) over the number of intervening words but are still well below that attained by simple edge counting.

---

[4] Words/Edge is a metric of density analogous to People/Square Mile.

|  | Human | | WordNet | | Roget's | | | |
|---|---|---|---|---|---|---|---|---|
| Word-Pair | Miller and Charles | Resnik | Information Content | Edges | Information Content | Edges | Intervening Words | Jiang & Conrath |
| car-automobile | 3.92 | 3.90 | 8.04 | 30 | 10.77 | 0 | 5 | 10.68 |
| gem-jewel | 3.84 | 3.50 | 14.93 | 30 | 13.23 | 0 | 1 | 12.47 |
| journey-voyage | 3.84 | 3.50 | 6.75 | 29 | 8.90 | 2 | 14 | 8.89 |
| boy-lad | 3.76 | 3.50 | 8.42 | 29 | 12.91 | 0 | 1 | 12.30 |
| coast-shore | 3.70 | 3.50 | 10.81 | 29 | 11.61 | 0 | 1 | 11.40 |
| asylum-madhouse | 3.61 | 3.60 | 16.67 | 29 | 11.16 | 0 | 2 | 11.04 |
| magician-wizard | 3.50 | 3.50 | 13.67 | 30 | 4.75 | 4 | 17 | 4.75 |
| midday-noon | 3.42 | 3.60 | 12.39 | 30 | 15.77 | 0 | 2 | 13.12 |
| furnace-stove | 3.11 | 2.60 | 1.71 | 23 | 13.53 | 0 | 1 | 12.66 |
| food-fruit | 3.08 | 2.10 | 5.01 | 27 | 0.02 | 4 | 369 | 0.02 |
| bird-cock | 3.05 | 2.20 | 9.31 | 29 | 1.47 | 4 | 47 | 1.47 |
| bird-crane | 2.97 | 2.10 | 9.31 | 27 | 1.47 | 4 | 919 | 1.47 |
| tool-implement | 2.95 | 3.40 | 6.08 | 29 | 13.35 | 0 | 1 | 12.54 |
| brother-monk | 2.82 | 2.40 | 2.97 | 24 | 9.89 | 2 | 2 | 9.85 |
| crane-implement | 1.68 | 0.30 | 2.97 | 24 | 2.53 | 4 | 336 | 2.53 |
| lad-brother | 1.66 | 1.20 | 2.94 | 26 | 0.00 | 10 | 15418 | 0.00 |
| journey-car | 1.16 | 0.70 | 0.00 | 0 | 0.84 | 6 | 478 | 0.84 |
| monk-oracle | 1.10 | 0.80 | 2.97 | 24 | 0.00 | 12 | 12052 | 0.00 |
| food-rooster | 0.89 | 1.10 | 1.01 | 18 | 0.00 | 12 | 25339 | 0.00 |
| coast-hill | 0.87 | 0.70 | 6.23 | 26 | 0.00 | 10 | 14024 | 1.91 |
| forest-graveyard | 0.84 | 0.60 | 0.00 | 0 | 0.30 | 8 | 296 | 0.30 |
| monk-slave | 0.55 | 0.70 | 2.97 | 27 | 0.00 | 12 | 29319 | 0.00 |
| coast-forest | 0.42 | 0.60 | 0.00 | 0 | 0.00 | 10 | 4801 | 1.91 |
| lad-wizard | 0.42 | 0.70 | 2.97 | 26 | 0.00 | 12 | 64057 | 0.00 |
| chord-smile | 0.13 | 0.10 | 2.35 | 20 | 0.00 | 8 | 3253 | 0.00 |
| glass-magician | 0.11 | 0.10 | 1.01 | 22 | 0.00 | 12 | 82965 | 0.00 |
| noon-string | 0.08 | 0.00 | 0.00 | 0 | 1.58 | 6 | 779 | 1.58 |
| rooster-voyage | 0.08 | 0.00 | 0.00 | 0 | 0.00 | 12 | 34780 | 0.00 |

**Table 1. Metric Results**

## Conclusion

This paper presented the results of using Roget's International Thesaurus as the taxonomy in a semantic similarity measurement task. Four similarity metrics were taken from the literature and applied to Roget's. The experimental evaluation suggests that the traditional edge counting approach does surprisingly well (a correlation of r=0.8862 with a benchmark set of human similarity judgements, with an upper bound of r=0.9015 for human subjects performing the same task.)

The results should provide incentive to those wishing to understand the effect of various attributes on metrics for semantic relatedness across hierarchies. Further investigation of why this dramatic improvement in edge counting occurs in the shallow, uniform hierarchy of Roget's needs to be conducted. The results should prove beneficial to those doing research with Roget's, WordNet and other semantic based hierarchies.

**Acknowledgements**

This research was sponsored in part by AFOSR under RL-2300C601.